\documentclass[a4paper, 10pt, conference]{ieeeconf}      % Use this line for a4
\IEEEoverridecommandlockouts                              % This command is only
\usepackage{graphics} % for pdf, bitmapped graphics files
\usepackage{graphicx}
\usepackage{xcolor}
\usepackage{amsmath}
\usepackage{textcomp}
\graphicspath{{calibrationimages/}}
\DeclareMathOperator*{\argmin}{argmin}
\title{\LARGE \bf
Automatic extrinsic calibration between a camera and a 3D Lidar using 3D point and plane correspondences 
}

\author{Surabhi Verma$^{1}$, Julie Stephany Berrio$^{1}$, Stewart Worrall$^{1}$,   Eduardo Nebot$^{1}$% <-this % stops a space
\thanks{*This work was not supported by any organization}% <-this % stops a space
\thanks{$^{1}$S. Verma, J. Berrio, S. Worrall, E. Nebot  are with the Australian Centre for Field Robotics (ACFR) at the University of Sydney (NSW, Australia).
       E-mails: {\tt\small \{s.verma, j.berrio, s.worrall, e.nebot\}@acfr.usyd.edu.au}}% <-this % stops a space
}

\begin{document}

\maketitle
\thispagestyle{empty}
\pagestyle{empty}

%%%%%%%%%%%%%%%%%%%%%%%%%%%%%%%%%%%%%%%%%%%%%%%%%%%%%%%%%%%%%%%%%%%%%%%%%%%%%%%%
\begin{abstract}

This paper proposes an automated method to obtain the extrinsic calibration parameters between a camera and a 3D lidar with as low as $16$ beams. We use a checkerboard as a reference to obtain features of interest in both sensor frames. The calibration board centre point and normal vector are automatically extracted from the lidar point cloud by exploiting the geometry of the board. The corresponding features in the camera image are obtained from the camera's extrinsic matrix. We explain the reasons behind selecting these features, and why they are more robust compared to other possibilities. To obtain the optimal extrinsic parameters, we choose a genetic algorithm to address the highly non-linear state space. The process is automated after defining the bounds of the 3D experimental region relative to the lidar, and the true board dimensions. In addition, the camera is assumed to be intrinsically calibrated. Our method requires a minimum of $3$ checkerboard poses, and the calibration accuracy is demonstrated by evaluating our algorithm using real world and simulated features.
\end{abstract}

%%%%%%%%%%%%%%%%%%%%%%%%%%%%%%%%%%%%%%%%%%%%%%%%%%%%%%%%%%%%%%%%%%%%%%%%%%%%%%%%
\section{INTRODUCTION}

Autonomous systems, using a multitude of sensors including 3D lidar and cameras, are being increasingly used for research and industrial applications. To enable higher levels of autonomy however, we need to extend the capabilities of such robots in order for them to construct accurate and complete models of their environment. One way to achieve this is by combining complementary information provided by the camera and lidar. Cameras, while being capable of supplying dense information in terms of color, texture, and shape of objects, are limited in their ability to provide high quality depth information at longer ranges. Lidars, on the other hand, capture accurate depth information with the drawback of lower vertical spatial resolution and a lack of colour/texture information. Furthermore, lidars are more suited to environments with variation in illumination and weather conditions. The fusion of camera and lidar makes it possible to overcome the limitations of each individual sensor. The main challenge in fusing these two different sensor modalities is the requirement for a precise calibration of the camera's intrinsic parameters, and the geometrical extrinsic parameters which includes the 3D transformation between the two sensors \cite{IROS_Berrio}.

Given accurate extrinsic parameters, it is possible to transform the lidar point cloud to the camera frame and then project these points on an image based on the intrinsic parameters for the camera. Such a projection can be used for various purposes, for example in [2], segmented point clouds are used as the ground truth to validate the performance of a CNN for a specific label. These parameters also enable camera to laser projection, where the corresponding image pixel data (RGB values, labels, etc.) is used to augment the point cloud fields. For example, authors in [3] translate the semantic information from images to point clouds to generate 3D semantic maps. Applications such as these can improve the overall perception of the robot. We therefore propose an approach to obtain the extrinsic parameters, describing the relative 3D rotation and translation, between a camera and a 3D lidar. This is a particularly challenging problem as the object features are obtained from different sensors
with different modalities and noise patterns. Noisy features reduce the accuracy of calibration. Furthermore, not all lidars/cameras have similar behavior and measurement errors making it difficult to generalize an approach. We address these issues by selecting features that are less susceptible to noise from sensor measurements, and by using a robust optimization strategy. It is challenging to calibrate a lidar sensor with a small number of beams as this results in a reduced vertical resolution. Our approach is designed to provide an accurate calibration for lidars with as low as 16-beams, though it is equally applicable to lidars with more beams.

Previously, several approaches have been proposed to solve the camera-lidar calibration problem. These approaches can be broadly classified into two categories: target-based and target-less methods. Targets such as a checkerboard, fiducial marker \cite{mirzaei} \cite{dhall} or custom-made target \cite{circular_target} \cite{arbitrary_trihed} have been used to find correspondences between features perceived in both the sensor frames. On the other hand, a target-less method uses features from natural scenes. Some of these methods \cite{Scaramuzza2007ExtrinsicSC} involve feature correspondences in the image and point cloud that are input manually. These approaches tend to require a significant number of features in the environment to reach a satisfactory calibration accuracy. Our proposed strategy uses a planar checkerboard target to compute the calibration parameters. With this approach, the inner corners of a checkerboard can be detected by the camera with sub-pixel accuracy and a planar model can be fit to the set of lidar points corresponding to the board. We will show that the automated extraction of lidar features leads to an accurate estimate of the board's geometric features. Such a method is appealing for automotive applications as the information required can be automatically obtained with minimal infrastructure within the vehicle production line.   

When it comes to choosing features to form geometric constraints, several possibilities arise. In our case, with a 3D lidar, we can extract the board normal, its distance from the lidar sensor, 3D lines (edges and lines on the board plane), and points (corner and centre). These features can also be found in the image. We specifically choose the checkerboard's normal and centre point to determine its orientation and 3D location respectively. Although lines and points on the board edges as described in \cite{Zhou-2018-107715} could generate sufficient constraints to solve the calibration problem with even a single checkerboard pose, we avoided using these features. This is because estimating board edges from the lidar data incorporates two sources of error: fitting a plane to the board points, and fitting a line to the edge points. As a result, the edges tend to have a higher error when compared to the board normal, which incorporates only a single source of error from the plane fitting. In addition, the edge points also suffer from beam divergence, a characteristic inherent to the lidar \textbf{\cite{velodyne}}. As a result, the edge lengths, found after fitting lines to the edge points, deviate from the ground truth by a few centimeters. Issues such as these make them less suitable for calibration. The centre point on the other hand, close to the centroid of the point cluster, is less prone to errors as points near the edges vary.

Once the feature correspondences are determined, a transformation matrix can be estimated between the camera and the lidar. For this, it is crucial to use an appropriate optimization algorithm as the geometric constraints used in the cost function tends to make it non-linear. Unlike other approaches that use gradient based algorithms \cite{Zhou-2018-107715} \cite{Zhang} \cite{inproceedings} \cite{zhou_part2} \cite{dong} which are susceptible to a local minima, we use a Genetic Algorithm (GA) which is a gradient-free method, similar to  \cite{Taylor2012AutomaticCO}.

In the next section, we present prior work relating to target based calibration with a checkerboard. In section III, we explain the details of automatically extracting the features from a planar checkerboard target and the optimization strategy. The experiments and outcomes are presented in section IV.
\section{RELATED WORK}

A checkerboard was first used by Zhang and Pless \cite{Zhang} to find the extrinsic parameters between a camera and a 2D Laser Rangefinder (LRF). In their method, for a number of checkerboard poses the checkerboard plane parameters are found relative to the camera. They then optimize for the transformation by minimizing the euclidean distance error between the laser points (after transformation) and the checkerboard plane (points on plane constraint). A similar approach was adopted by Unnikrishnan et al. \cite{article} to calibrate a 3D laser scanner and a camera. They manually choose the 2D region of interest in the laser range image to find plane correspondences in both sensor frames. A two stage optimization process is used which involves estimating the rotation and translation independently and then jointly optimizing the two sets of parameters. Pandey et al. \cite{inproceedings} calibrate an omni-directional camera and a 3D laser using the same geometric constraint as \cite{Zhang}. The issue with this approach is that the mere inclusion of the plane parameters, i.e. the distance of the checkerboard plane from the camera and its normal vector, to form a single constraint does not affix the location of the checkerboard on the plane. They may therefore require a higher number of checkerboard poses to converge to an optimal solution. In contrast, Zhou \cite{zhou_better_zhang} exploits two geometric constraints from plane-line correspondences to calibrate a 2D LRF and a camera. In \cite{dong}, the authors use a V-shaped calibration target formed by two triangular boards with a checkerboard on each triangle. As their target is non-planar, they are able to exploit the geometry of the setup to formulate a well-constrained cost function, minimizing point to plane distances. Our method uses a planar checkerboard, which can be used for the calibration of a camera as well. To overcome the need of obtaining several checkerboard poses for a satisfactory calibration, Geiger et al. \cite{ICRA12} attach multiple checkerboards in different scene locations. They are thus able to get their results with a single scene shot. In practice, it is not feasible to attach several checkerboards with accuracy every time the sensors are required to be calibrated. Zhou et al. \cite{zhou_part2} estimate the rotation and translation separately. Additionally, they introduce a weighting factor for each checkerboard pose depending on the measurement uncertainty from images. In order to decouple rotation and translation, features close to the ground truth are required, which is very difficult to obtain considering the noise in the sensors, particularly the lidar.

More recently in \cite{Zhou-2018-107715}, points corresponding to the edges of the checkerboard were used to fit lines. The corresponding edge lines were obtained from the camera and constraints were formed using plane-line correspondence. As explained in the previous section, these features are more prone to errors and hence we avoid using them. 

\section{METHODOLOGY}
Our method uses a checkerboard as a reference to obtain features of interest in the image and point cloud. For a given $i^{th}$ sample among $N$ camera$_{c}$-lidar$_{l}$ scan pairs of the checkerboard, we extract the centre point $(\mathbf{o}^i_\mathbf{{c/l}})$ and the normal vector $(\mathbf{n}^i_\mathbf{{c/l}})$ of the board. While the normal vector helps us determine the board plane's orientation, the centre point affixes the board's 3D location in each sensor's reference frame. Once the features are extracted, we exploit the correspondences, 
$\mathbf{o}^i_\mathbf{c} \leftrightarrow   \mathbf{o}^i_\mathbf{l}$ and $\mathbf{n}^i_\mathbf{c} \leftrightarrow   \mathbf{n}^i_\mathbf{l}$, to form constraints for rotation $(\mathbf{R^c_l})$ and translation $(\mathbf{t^c_l})$ of the lidar frame with respect to the camera frame.

The parameters required before starting the calibration process are list below:
\begin{itemize}
\item $(l, w)$: length and width of the rectangular board in metres on which the checkerboard is attached.
\item $(m, n)$: grid size of the checkerboard, i.e. the number of internal corners in each row and column. 
\item $(s)$: side length of each square within the checkerboard in metres.
\item $(bb_{x_{mn,mx}}, bb_{y_{mn,mx}}, bb_{z_{mn,mx}})$: minimum and maximum bounds of the 3D experimental region along the lidar's $x,y,z$ axis.
\item $(f_{xy}, c_{xy}, d_{1,2,3,4,5})$: camera intrinsic parameters including the focal length, principal point and 4/5 distortion coefficients corresponding to a fisheye/pinhole camera model. 
\end{itemize}

\subsection{Data collection setup}

To collect the samples required for calibration, we firmly attach a checkerboard on a rigid, opaque, and rectangular board such that both their centres align and their edges remain parallel to one another. In every sample, we ensure that the entire board is visible to both sensors and the 3D experimental region is free from any other objects apart from the board and its stand. The stand is chosen such that it does not hold the board with significant protruding elements close to the board boundaries or corners. This is necessary to conveniently filter out the points corresponding to the board. 
\begin{figure}[!h]
\centerline{
\includegraphics[width=0.48\textwidth]{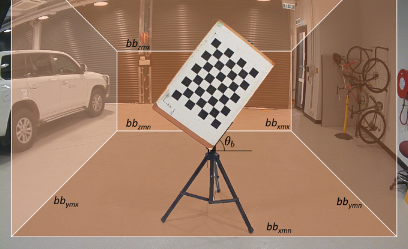}
}
\caption{\small Data collection setup. A virtual bounding box shows the experimental region inside which the board is located to obtain the samples. }
\label{fig:experimental_area}
\end{figure}

The board is kept tilted at an angle $\theta_b$ of around $45$ to $60$ degrees with respect to the ground plane. Such a configuration is adopted to offset the low vertical angular resolution of the lidar. Furthermore, we ensure that the checkerboard pattern is detectable by the camera and a minimum of $2$ lidar scan beams pass though each of the board's edges to allow the interpolation of points lying on the edges. We note that scan lines represent the board more accurately when the board is kept facing the sensor pair. Lastly, a diverse sample set is collected in terms of the board location in the experimental region. The data collection setup is shown in Fig. \ref{fig:experimental_area}.

\subsection{Feature Extraction}

\subsubsection{From the lidar} 

The minimum and maximum bounds ($bb_{x_{mn,mx}}, bb_{y_{mn,mx}}, bb_{z_{mn,mx}}$) allow us to separate the experimental region, consisting of the board and it's stand, from the environment point cloud. Since the stand has no protruding elements through which the board is held, we can remove the points corresponding to the legs of the stand in order to obtain the board point cloud. For this, the point with the maximum $z$-coordinate value is found in the experimental region. As this point lies close the board's top most corner in the tilted configuration, we reset $bb_{z_{mx}}$ to this $z$-coordinate value and accordingly redefine $bb_{z_{mn}}$ to extract the board cloud.

\begin{equation} \label{eq1}
bb_{z_{mn}}=bb_{z_{mx}} - \sqrt{l^2_b + w^2_b}
\end{equation}

(\ref{eq1}) was formulated assuming that the difference in angle between the board plane and the $y-z$ plane of the lidar frame is small enough to obtain the lowermost point of the board by subtracting the newly defined $bb_{z_{mx}}$ with the board diagonal.  

\begin{figure}[!h]
\centerline{
\includegraphics[width=0.48\textwidth]{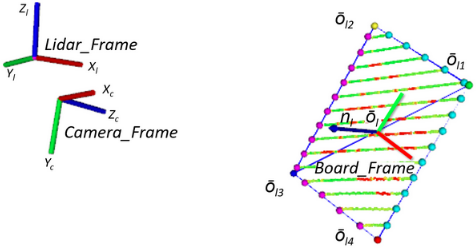}
}
\caption{\small Sensor frames and features extracted from the point cloud.}
\label{fig:pc_features}
\end{figure}

Next, we fit a plane to the filtered cluster of board points using 3D RANSAC \cite{Fischler_Ransac} to get a robust estimate of the board plane. The obtained plane equation allows us to calculate $\mathbf{n}^i_\mathbf{l}$. In order to determine the centre point, we project the board point cloud to the estimated plane and determine the edge points by finding the points with the minimum and maximum $y$-coordinate value in each horizontal scan beam. The corner points of the board are then obtained by fitting $4$ lines through these edge points and calculating their intersection $\mathbf{\overline{o}^i_{l{k}}
}$; $k=1,2,3,4$. 
The line joining opposite corner points, $\mathbf{\overline{o}}^i_\mathbf{{l{1}}}$ and $\mathbf{\overline{o}}^i_\mathbf{{l{3}}}$, correspond to one of the board diagonals. The midpoint of this diagonal is $\mathbf{o}^i_\mathbf{l}$. Fig. \ref{fig:pc_features} depicts the sensor frames and the features extracted from the board point cloud. One may note that using the board edges and corner points, claimed to be noisy in the previous section, to find $\mathbf{o}^i_\mathbf{l}$ would make it less precise as well. However, the centre point remains immune to changes in the orientation of board due to lidar measurement errors and therefore to errors in the edges and corner points. 

\subsubsection{From the camera} 

\begin{figure}[!t]
\vspace{3mm}
\centerline{
\includegraphics[width=0.4\textwidth]{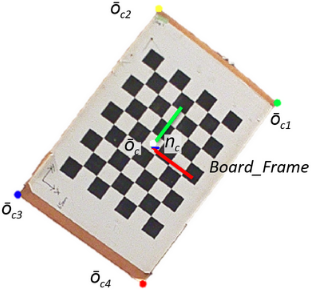}
}
\caption{\small Image extracted corner and centre points features.}
\label{fig:im_features}
\end{figure}

The checkerboard pattern in the image is detected using the OpenCV function cv::findChessboardCorners \cite{itseez2015opencv}. This information allows us to compute the pose of the board reference frame relative to the camera with the Perspective-n-Point (PnP) algorithm \cite{pnp}. In case of a pinhole camera, we can directly use the chessboard corners in the image frame and the board frame along with the camera intrinsic parameters as an input to PnP. For a fisheye camera model, we first undistort the the raw image and obtain the corresponding checkerboard corners. This is then fed to the PnP algorithm along with zero distortion coefficients. The output of PnP is the pose, rotation and translation, of the board reference frame (placed at the centre of the checkerboard) relative to the camera frame. We thereby obtain $\mathbf{o}^i_\mathbf{c}$. The $3^{rd}$ column of the rotation matrix, parametrized by the Euler angles, represents the z-axis of the board reference. This coincides with $\mathbf{n}^i_\mathbf{c}$ given that the $x,y$-axis of the board frame lies on the board plane and the $z$-axis, normal to it. Furthermore, we can find the corner points relative to the board centre knowing $l$ and $w$. This information along with the camera extrinsics, allow us to compute $\mathbf{\overline{o}}^i_\mathbf{{c{1,2,3,4}}}$ in the camera frame. The board features, as seen by the camera in its image is shown in Fig. \ref{fig:im_features}.

\subsection{Optimization Strategy}

After obtaining the required features, we optimize for the transformation from the lidar to camera frame. For this, we choose a Genetic Algorithm (GA) \cite{Mitchell:1998:IGA:522098} as the preferred optimizer. Due to it's stochastic nature, GA ``evolves'' towards better solutions without being susceptible to pitfalls of convergence to a local minima in our highly non-convex state space. Our optimization variables are the Euler angles $\boldsymbol{\theta^c_l} = [\theta_x, \theta_y, \theta_z]$ in the $xyz$-convention, and the translation $\mathbf{t^c_l} = [x$, $y$, $z]$ from the lidar to camera frame. To restrict the search space for these variables, we obtain an initial estimate ($\mathbf{\Tilde{R}^c_l}(\boldsymbol{\Tilde{\theta}^c_l})$ , $\mathbf{\Tilde{t}^c_l}$) of the required rotation and translation as follows. 

\begin{align}
\mathbf{\Tilde{R}^c_lN_l} & = \mathbf{N_c} \nonumber \\
\mathbf{N}^T_\mathbf{l}(\mathbf{\Tilde{R}^c_l})^T  & = \mathbf{N}^T_\mathbf{c} \nonumber \\
\mathbf{N_lN}^T_\mathbf{l}(\mathbf{\Tilde{R}^c_l})^T & = \mathbf{N_lN}^T_\mathbf{c} \nonumber \\
(\mathbf{\Tilde{R}^c_l})^T & = (\mathbf{N_lN}^T_\mathbf{l})^{-1}(\mathbf{N_lN}^T_\mathbf{c}) \nonumber \\
\mathbf{\Tilde{R}^c_l} & =((\mathbf{N_lN}^T_\mathbf{l})^{-1}(\mathbf{N_lN}^T_\mathbf{c}))^T 
\end{align}

The notations we use are, $\mathbf{N_l, N_c, O_l}$ and $\mathbf{O_c}$, representing $3$x$N$ matrices comprising of the column vectors $\mathbf{n^i_l, n^i_c, o^i_l}$, and $\mathbf{o^i_c}$, for $i= 1, 2, ..., N$ samples, respectively. To refine $\mathbf{\Tilde{R}^c_l}(\boldsymbol{\Tilde{\theta}^c_l})$, we first optimize for rotation before jointly optimizing for $\mathbf{R^c_l}(\boldsymbol{\theta^c_l})$ and $\mathbf{t^c_l}$. The components used to form the fitness function for the rotation optimization are:

\begin{enumerate}

\item The dot product average between the vector $(\mathbf{o}^{i}_\mathbf{c}$ - $\mathbf{\overline{o}}^{i}_\mathbf{c1})$ and $\mathbf{n}^{i}_\mathbf{l,c}$ (this value should ideally be $0$ if the transformed normal $\mathbf{n}^{i}_\mathbf{l,c}$ is perpendicular to the vector $(\mathbf{o}^{i}_\mathbf{c}$ - $\mathbf{\overline{o}}^{i}_\mathbf{c1})$ lying on the plane) 
\begin{equation} \label{eq:subeq5}
    e_{d} =\frac{1}{N}\left \{  \sum_{i=1}^{N}\left (\left(\mathbf{o}^{i}_\mathbf{c}\! - \mathbf{\overline{o}}^{i}_\mathbf{c1}\right)\cdot\mathbf{n}^i_\mathbf{{l,c}}\right )^2\right \}
\end{equation}
where, $\mathbf{n}^{i}_\mathbf{l,c}$ (=$\mathbf{R^c_l}\mathbf{n}^i_\mathbf{l}$) is $\mathbf{n}^i_\mathbf{l}$ in the camera frame. 
\item The alignment between $\mathbf{n}^{i}_\mathbf{l,c}$ and $\mathbf{n}^{i}_\mathbf{c}$ 
\begin{equation} \label{eq:subeq4}
    e_{r} =\frac{1}{N}\left \{  \sum_{i=1}^{N}\left (\sqrt{\sum_{}^{} \left (\mathbf{n}^{i}_\mathbf{l,c}\! - \mathbf{n}^{i}_\mathbf{c}  \right )^2}\right )\right \}
\end{equation}
\end{enumerate}
The fitness function combining the above factors is:
\begin{subequations}
\begin{equation}
\label{eq:subeq5a}
\begin{split}
    \mathbf{{\Tilde{R}}^c_l}(\boldsymbol{{\Tilde{\theta}}^c_l}) = \argmin_{\mathbf{{R}^c_l}(\boldsymbol{{\theta}^c_l})} e_d + e_r
\end{split}
\end{equation}
Once we get a good estimate of $\mathbf{{\Tilde{R}}^c_l}$ in terms of lower values of $e_r$ and $e_d$, we obtain $\mathbf{{\Tilde{t}}^c_l}$; the translation estimate,
\begin{equation}
\label{eq:subeq5b}
\mathbf{{\Tilde{t}}^c_l}= mean(\mathbf{O_c} - \mathbf{{\Tilde{R}}^c_l}\mathbf{O_l}),
\end{equation}
\end{subequations}
where $mean$ represents the average operation performed row-wise. Knowing, ($\mathbf{{\Tilde{R}}^c_l}$, $\mathbf{{\Tilde{t}}^c_l}$), we constrict the bounds of the joint search space for $\mathbf{{R}^c_l}$ and $\mathbf{{t}^c_l}$. We form the corresponding fitness function by introducing the following factors in addition to $e_d$ and $e_r$. 

\begin{enumerate}

\item The euclidean distance average between $\mathbf{o}^{i}_\mathbf{l,c}$ and $\mathbf{o}^{i}_\mathbf{c}$,
\begin{equation} \label{eq:subeq7}
    e_{t} =\frac{1}{N}\left \{  \sum_{i=1}^{N}\left (\sqrt{\sum_{}^{} \left (\mathbf{o}^{i}_\mathbf{c}\! - \mathbf{o}^{i}_\mathbf{l,c}  \right )^2}\right )\right \},
\end{equation}
where, $\mathbf{o}^{i}_\mathbf{l,c}$ (=$\mathbf{R^c_l}\mathbf{o}^{i}_\mathbf{l} + \mathbf{{t}^c_l}$) is $\mathbf{o}^{i}_\mathbf{l}$ measured in the camera frame.

\item The variance in the euclidean distance between $\mathbf{o}^{i}_\mathbf{{l,c}}$ and $\mathbf{o}^{i}_\mathbf{c}$, in all the samples $N$. 
\begin{equation} \label{eq:subeq8}
v_{t} = \frac{1}{N}\left \{  \sum_{i=1}^{N}\left (\sqrt{\sum_{}^{}\left (\mathbf{o}^{i}_\mathbf{c}\! - \mathbf{o}^{i}_\mathbf{l,c}  \right )^2} - e_{t}\right )^2\right \}
\end{equation}

\end{enumerate}
The variance component is included to avoid any bias in the euclidean distance for a particular sample.  Note: The units of each of the above  components  are  in  metres,  same  as  that  of  all  the prerequisite measurements in the calibration process. The range of distance errors in metres are similar to that of rotation errors in radians, eliminating the need for normalizing any variable. 

The factors defined in (\ref{eq:subeq7}) and (\ref{eq:subeq8}), consider the errors and variance in 3D Cartesian space. To incorporate the errors in the 2D image plane, we minimize the maximum value of the re-projection error between the centre points, $\mathbf{o}^{i}_\mathbf{{l,c}}$ and $\mathbf{o}^{i}_\mathbf{c}$, amongst all the samples: 
\begin{equation} 
    e_{t,I} =  max\left \{\sqrt{\sum_{}^{}\left (\mathbf{o}^{i}_\mathbf{c,I}\! - \mathbf{o}^{i}_\mathbf{l,c,I}  \right )^2}\right \}
\end{equation}
such that, $i=1,2,..,N$. Since the error in a 2D image is in pixels, we convert it into metres in order to be able to add them with the errors obtained previously. This is done by finding the number of pixels $p_l$ lying on the edge of the square located in the middle of the checkerboard. Knowing $s$, we can then find the metre length corresponding to 1 pixel for a particular distance of the checkerboard in the $i^{th}$ sample. We assume this metre correspondence of a pixel to be constant for all the pixels lying close to the board centre. This conversion is hence applied to  $e_{t,I}$. The combined fitness function is defined below. 
\begin{equation} \label{eq:subeq10}
    (\mathbf{{\hat{R}}^c_l}(\boldsymbol{{\hat{\theta}}^c_l}),\mathbf{\hat{t}^c_l}) = \argmin_{\mathbf{{R}^c_l}(\boldsymbol{{\theta}^c_l}), \mathbf{t^c_l}} e_{t} + v_{t} + e_d + e_r + ke_{t,I},
\end{equation}
subject to: 
\begin{align}
    \boldsymbol{{\theta}^c_l} \in \boldsymbol{\Tilde{\theta}^c_l} \pm \pi/18 \nonumber\\
    \mathbf{{t}^c_l} \in \mathbf{{\Tilde{t}}^c_l} \pm 0.05 \nonumber
\end{align}
In (\ref{eq:subeq10}), $k (= \frac{s}{p_l})$ is the error conversion from pixel to metre and ($\boldsymbol{{\theta}^c_l}, \mathbf{{t}^c_l}$) is bound to vary within $\pm$ $\pi/18$ radians or $10$ degrees from $\boldsymbol{{\Tilde{\theta}}^c_l}$ and $0.05$ metres from $\mathbf{\Tilde{t}^c_l}$ respectively. As ($\boldsymbol{{\Tilde{\theta}}^c_l}$, $\mathbf{\Tilde{t}^c_l}$) have been refined in (\ref{eq:subeq5a}) and (\ref{eq:subeq5b}), we can set a small bound.

\section{EXPERIMENTAL RESULTS}

To test the robustness of our algorithm, we conduct experiments with simulated and real data. Furthermore, we verify our results by visually inspecting the projection of the point cloud on the image with two different sensor configurations. 

For all the experiments, we initiate GA with a population size of 200 lying within the variable bounds of (\ref{eq:subeq10}). Due to the non-deterministic nature of GA, we run the process 10 times for a given sample set, $N$, and take the average of the obtained extrinsic parameters. The average value is considered as the final extrinsics obtained from that particular $N$. Our algorithm took between 2-10 minutes for $N$ ranging from 3-30.

\subsection{Simulated Data}
In order to perform the calibration in a simulated scenario, we generate the sensor frames with an arbitrary transform and place the board (board centre) in different possible locations in 3D space. This is done making sure that the board is visible to both sensors. Next, we vary the orientation of the board around a randomly chosen axis. Since we start with a known transform of both sensor frames relative to the world frame, we can find the features of interest as measured by both sensors. However, sensor measurements (especially the lidar) have errors involved in them. Due to this, the lidar data points from a static scene differ in each time step. By observing this variation, we note that the range of the board normal and board centre, results to around $4^\circ$ and $1$ cm respectively. This forms the basis for generating the noise in the simulated features as obtained from the lidar. We add three different noise levels to the board normal, $\pm$ $1.5^\circ$, $\pm$ $2^\circ$, and $\pm$ $2.5^\circ$. The normal vectors have a Gaussian distribution with the ground truth normal as the mean, deviating up to the noise level. The centre point is displaced within a sphere of $1$ cm diameter, with the ground truth as the sphere centre. This displacement has a Gaussian distribution across samples. We do not add any noise to the features obtained by the camera as we just require noisy data, be it from the camera or the lidar.
\begin{figure}[h]
\vspace{1mm}
\centerline{
\includegraphics[width=0.48\textwidth]{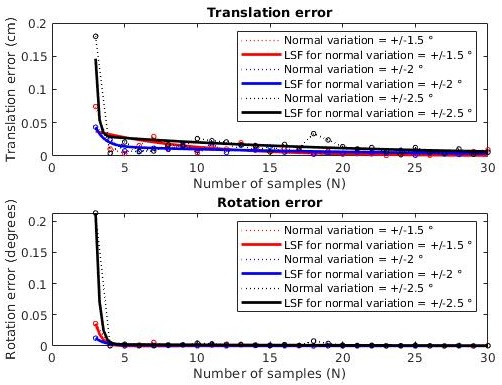}
}
\caption{\small Translation and rotation error for simulated data.}
\label{fig:simulated_spread}
\end{figure} 

We run the optimizer starting with an input of 3 random samples, the minimum number required by our algorithm. As the ground truth $(\mathbf{R^c_l}, \mathbf{t^c_l})$ and estimated transform $(\mathbf{\hat{R}^c_l}, \mathbf{\hat{t}^c_l})$ are known, we can calculate the translation and rotation error \cite{Huynh2009} as follows:
\begin{align}
e_{translation} & = \| \mathbf{t^c_l - \hat{t}^c_l} \| \\
e_{rotation} & = \| \mathbf{I} - \mathbf{(R^c_l)^{-1}\hat{R}^c_l} \|_F 
\end{align}
where, $\|.\|$ denotes the norm operation and $\|.\|_F$ is the Frobenius norm. Fig. \ref{fig:simulated_spread} shows the errors in rotation and translation as more samples get introduced into the optimizer. We note that some samples can be relatively more noisy when compared to the rest. This is due to the way the board is placed relative to the lidar, the number of scan lines passing through it and the corresponding plane model estimation errors. So depending upon the quality of the incoming samples, the rotation and translation errors might increase or decrease relative to the error from the previous sample set. However, the overall tendency of the error is to decay exponentially as more samples are added. This is represented by taking the Least Square Fit (LSF) through the data points. It can be noticed in Fig. \ref{fig:simulated_spread} that the red colored LSF curve, representing a lower noise of $\pm$ $1.5^\circ$ when compared to $\pm$ $2^\circ$ by the blue colored LSF curve, eventually surpasses the blue LSF curve and leads to a lower transformation error in comparison. The green LSF curve, corresponding to a $\pm$ $2.5^\circ$ noise shows a higher convergence error in comparison. Nonetheless, the error converges to less than $0.5$ cm for translation and close to $0$ degree error for rotation with a high sample number. After $N \approx 9$, the error tends to become constant across samples. Therefore, we do our analysis upto $9$ samples with the real data.

\subsection{Real Data}
\begin{figure}[h]
\vspace{1mm}
  \centering
  \begin{minipage}[b]{0.09\textwidth}
    \includegraphics[width=\textwidth]{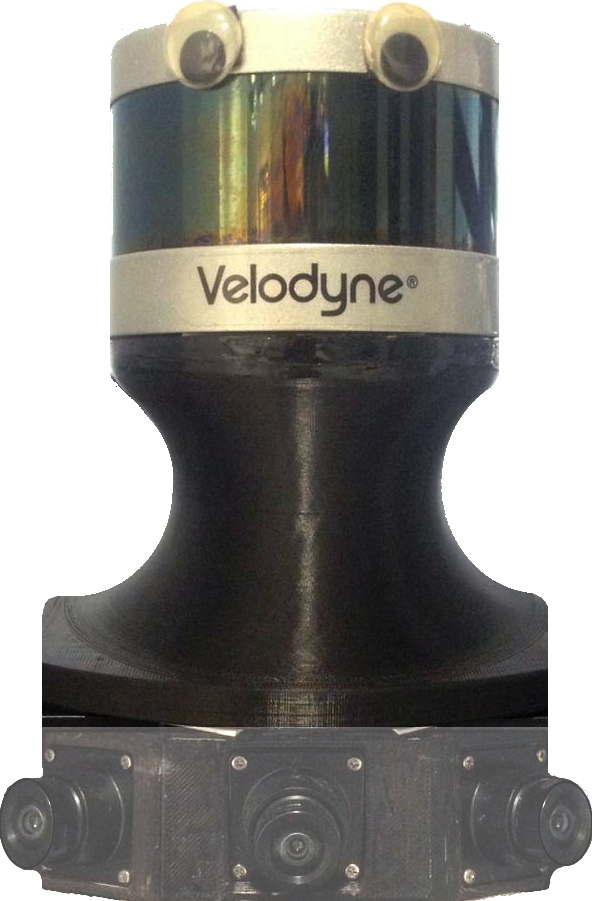}
    \caption{Sensor setup.}
    \label{fig:sensors}
  \end{minipage}
  \hfill
  \begin{minipage}[b]{0.35\textwidth}
    \includegraphics[width=\textwidth]{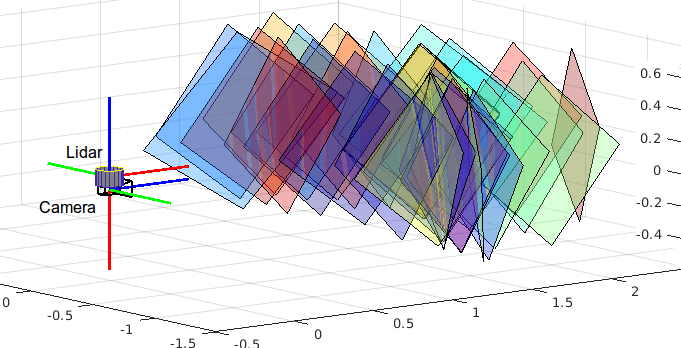}
    \caption{Real samples used to evaluate our calibration algorithm}
    \label{fig:samples}
  \end{minipage}
\end{figure}

We obtain real data, i.e. the point cloud corresponding to the board and the image of the checkerboard, from Velodyne's VLP-16 lidar and NVIDIA's 2Mega SF3322 automotive GMSL camera. The sensor setup for data collection is shown in Fig. \ref{fig:sensors} and we specifically calibrate the camera located in the centre and the lidar. VLP-16 is a $16$ beam lidar with a $360^\circ$ horizontal field of view (FOV) and $\pm 15^\circ$ vertical FOV. The range accuracy is $\pm3 cm$. The GMSL camera lens has a $100^\circ$ horizontal FOV and $60^\circ$ vertical FOV. Images can be captured at 30 frame per second (fps) with the resolution of $1928 \times 1208$ (2.3M pixel).  

\begin{figure}[h]
\vspace{3mm}
\centerline{
\includegraphics[width=0.475\textwidth]{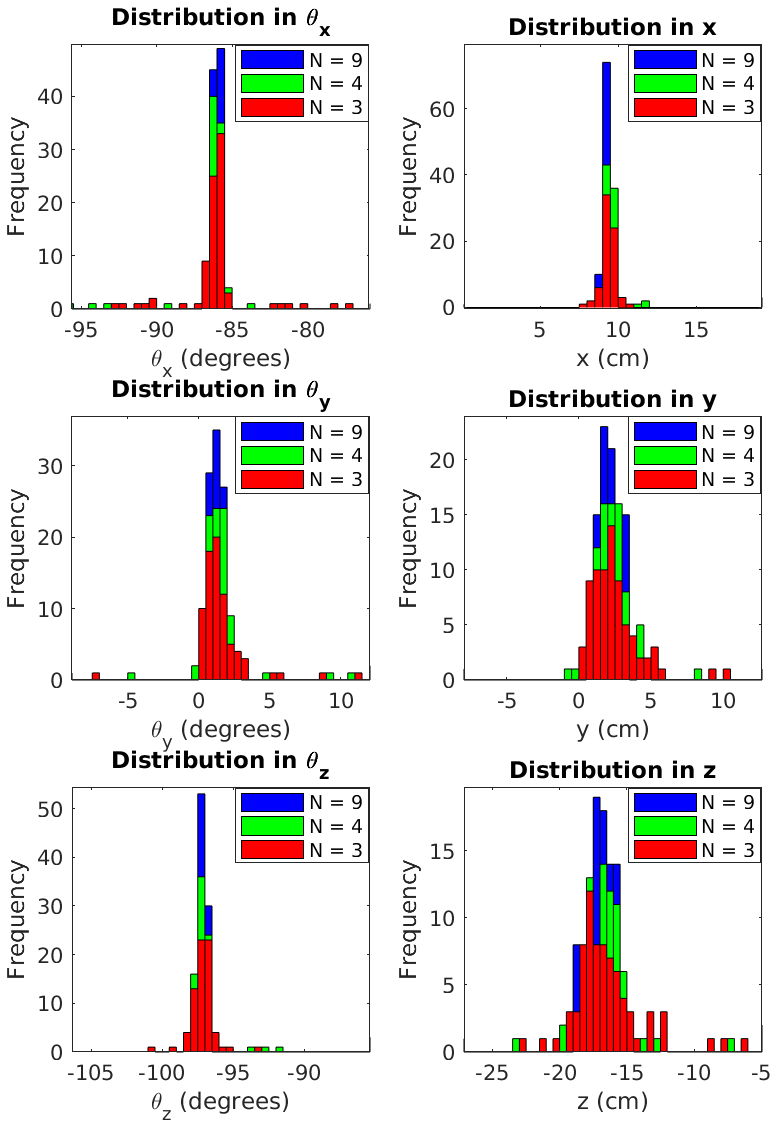}}
\caption{\small Histogram plots implying a low spread in the calibration parameters. }
\label{fig:spread}
\end{figure}

To evaluate our approach with real data, we collect a set of $30$ input samples as shown in Fig. \ref{fig:samples}. We begin by randomly choosing $100$ sets out of the different possible combinations of $3$ samples and feed them into the optimizer. Thereafter, we increment the samples for which the combinations are obtained. In Fig. \ref{fig:spread}, we show the distribution of $100$ randomly chosen extrinsic parameters obtained from the combinations of $3$, $4$ and $9$ samples. The width of each bin for translation and rotation is $0.5$ centimetre and $0.5^\circ$ respectively. It can be observed that as N increases, the spread in the extrinsic parameters decrease. This is indicative of the robustness of our algorithm across samples. The outliers in the histogram can be attributed to an improper/noisy sample set, which decreases as N is incremented.

\begin{figure}[h]
\vspace{1mm}
\centerline{
\includegraphics[width=0.48\textwidth]{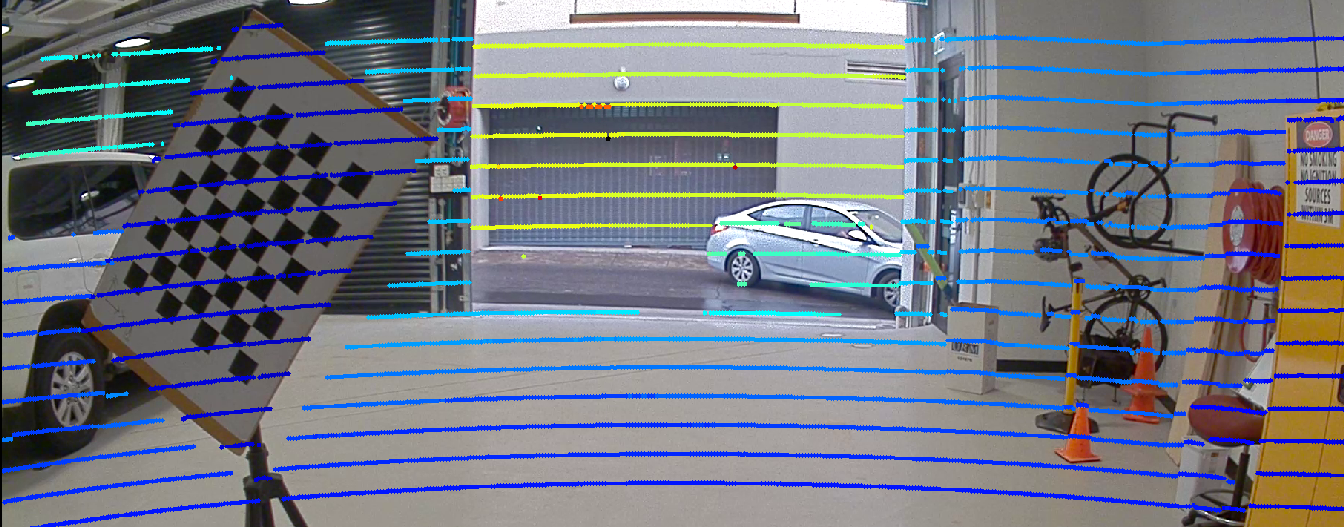}
}
\caption{\small Point cloud projection on an image for 9 samples. }
\label{fig:projection}
\end{figure}

\begin{figure}[h]
\vspace{3mm}
\centerline{
\includegraphics[width=0.48\textwidth]{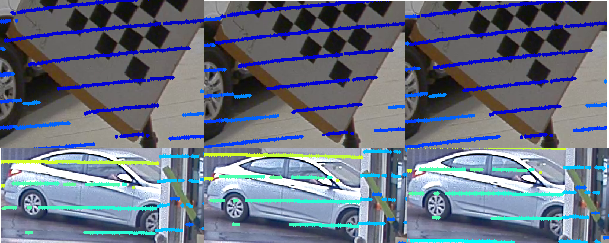}
}
\caption{\small Projection for N = 3, N = 4, N = 9. }
\label{fig:projection_diff}
\end{figure}

For a qualitative analysis, we project the point cloud on the image as shown in Fig. \ref{fig:projection}. In this image, the color of the points vary relative to the distance between the obstacle and the lidar. Assuming that the intrinsics are correct, an accurate extrinsic calibration would imply a projection such that there is a visually evident correspondence between the boundaries of objects in the point cloud and the edges of objects in the image. We specifically chose to visualize the projection of the laser points to the entire image and not just the board. This is because the highly non-linear fitness function can lead to an accurate projection of points on the board, i.e. near the sample point used in the optimizer, but not elsewhere in the image due to over-fitting. We can see this happening in Fig. \ref{fig:projection_diff}, where the board projection is visually accurate when $N = 3, 4$, and $9$, but the overall projection improves as $N$ increases. Also, the trend of the graph observed in Fig. \ref{fig:simulated_spread} is evident in the projection where there is significant improvement from $N = 3$ to $N = 4$ when compared $N = 4$ to $N = 9$.

When closely observing the calibration boards in Fig. \ref{fig:projection_diff}, we notice that the board lidar projection extends slightly beyond the board edges in each image. This can be attributed to the spot size of the laser points. According to the VLP-16 data sheet, the laser spot size at around $2$ metres, the distance from the lidar at which the board is kept, is $18.2$ millimetres in the horizontal direction. Assuming that the spot centre falls at the board edge, this is an error of around $9$ millimetres, equivalent to $6-7$ pixels, in the horizontal direction. We note that as this error is equal for each edge of the board, it does not affect the position of the board centre. This reinforces the appropriateness of choosing the centre point in comparison to other more noisy features such as the board corners or its edges.  

\section{CONCLUSIONS AND FUTURE WORK}

In this paper we presented a robust and automated approach to estimate the extrinsic calibration parameters between a pinhole/fisheye camera and 3D lidar using a planar checkerboard. For this, we chose the most stable features based on the errors in the lidar measurements (compared to other features), to obtain the 3D point and plane correspondences. Our method automatically extracted these features for calibration and GA was used to obtain a globally optimal calibration result.

We demonstrated experimentally that our method is able to obtain consistent results which improve as more samples are added into the optimizer. Occasional outliers can be attributed to the measurement error and quality of the sample. An analysis with the simulated data and the image projection made us conclude that a globally optimal solution can be achieved by adding additional samples (N = 9 or 10), beyond the minimum requirement of $3$, into the optimizer.   

In future, we plan to synchronize the sensor data and apply motion correction to the lidar scans so that the mobile platform can be moved around the calibration target to collect different samples and obtain the extrinsic parameters at run-time. This could be very useful in the production line for autonomous vehicles or other robots. 
%\addtolength{\textheight}{-5cm}   % This command serves to balance the column lengths

%%%%%%%%%%%%%%%%%%%%%%%%%%%%%%%%%%%%%%%%%%%%%%%%%%%%%%%%%%%%%%%%%%%%%%%%%%%%%%%%

%%%%%%%%%%%%%%%%%%%%%%%%%%%%%%%%%%%%%%%%%%%%%%%%%%%%%%%%%%%%%%%%%%%%%%%%%%%%%%%%

%%%%%%%%%%%%%%%%%%%%%%%%%%%%%%%%%%%%%%%%%%%%%%%%%%%%%%%%%%%%%%%%%%%%%%%%%%%%%%%%

\section*{ACKNOWLEDGMENT}

This work has been funded by the ACFR and the Australian Research Council Discovery Grant DP160104081 and University of Michigan / Ford Motors Company Contract ``Next generation Vehicles".

\bibliography{main}
\bibliographystyle{IEEEtran}

%\begin{thebibliography}{99}

%\end{thebibliography}

\end{document}